\documentclass[lettersize,journal]{IEEEtran}
\usepackage{amsmath,amsfonts}
\usepackage{array}
\usepackage[caption=false,font=normalsize,labelfont=sf,textfont=sf]{subfig}
\usepackage{textcomp}
\usepackage{stfloats}
\usepackage{url}
\usepackage{verbatim}
\usepackage{graphicx}
\usepackage{multirow} 
\usepackage{cite}
\usepackage{bm}
\usepackage{makecell}
\usepackage{booktabs} 
\usepackage{mathtools}
\usepackage{hyperref}

\hypersetup{
    colorlinks=true,              
    linkcolor=blue,               
    citecolor=blue,               
    urlcolor=blue,                
    filecolor=blue,               
}

\usepackage[linesnumbered,ruled,vlined]{algorithm2e}

\begin{document}

\title{SAFL: Structure-Aware Personalized Federated Learning via Client-Specific Clustering and SCSI-Guided Model Pruning}

\author{Nan Li, Xiaolu Wang, Xiao Du, Puyu Cai, Ting Wang
\thanks{Nan Li , Xiao Du and Ting Wang are with the Engineering Research Center of Software/Hardware Co-Design Technology and Application, Ministry of Education, and the Shangh
ai Key Laboratory of Trustworthy Computing, East China Normal University, Shanghai 200062, China (e-mail:
51265902071@stu.ecnu.edu.cn; 52265902007@stu.ecnu.edu.cn; twang@sei.ecnu.edu.cn).}
\thanks{Xiaolu Wang is with the Software Engineering Institute, East China Normal University, Shanghai 200050, China (e-mail: xiaoluwang@sei.ecnu.edu.cn).}
\thanks{Puyu Cai is with the Computer Science Department at New York University, New York, NY 10012, United States. (e-mail: pc3128@nyu.edu).}}




\maketitle
\begin{abstract}
Federated Learning (FL) enables clients to collaboratively train machine learning models without sharing local data, preserving privacy in diverse environments. While traditional FL approaches preserve privacy, they often struggle with high computational and communication overhead. To address these issues, model pruning is introduced as a strategy to streamline computations. However, existing pruning methods, when applied solely based on local data, often produce sub-models that inadequately reflect clients' specific tasks due to data insufficiency. To overcome these challenges, this paper introduces SAFL (Structure-Aware Federated Learning), a novel framework that enhances personalized federated learning through client-specific clustering and Similar Client Structure Information (SCSI)-guided model pruning. SAFL employs a two-stage process: initially, it groups clients based on data similarities and uses aggregated pruning criteria to guide the pruning process, facilitating the identification of optimal sub-models. Subsequently, clients train these pruned models and engage in server-based aggregation, ensuring tailored and efficient models for each client. This method significantly reduces computational overhead while improving inference accuracy. Extensive experiments demonstrate that SAFL markedly diminishes model size and improves performance, making it highly effective in federated environments characterized by heterogeneous data.
\end{abstract}


\section{Introduction}
\IEEEPARstart{F}{erated} learning (FL) has become a significant advancement in the realm of distributed machine learning, particularly in addressing privacy issues that are prevalent in traditional methods \cite{mcmahan2017communication}, \cite{konevcny2016federated}, \cite{hard2018federated}. 
Unlike conventional approaches, FL allows each participant to keep their data on their local devices, thus preventing the transfer of sensitive data to a central server---a process that could expose them to privacy risks.
In the standard FL framework, clients work together to develop a unified global model. This collaboration is coordinated by a central server. During each learning cycle, clients download the latest model from the server, improve it using their own datasets, and then upload the updated parameters or gradients back to the server. This cycle repeats until the model converges. 
At the end of the process, all clients possess the same trained model, ensuring consistency across the model while maintaining the privacy of individual datasets.

Despite its advantages, standard FL can struggle with data that is non-uniformly distributed across clients—a scenario referred to as non-independent and identically distributed (non-IID) data \cite{zhao2018federated}, \cite{kairouz2021advances}, \cite{li2020federated}.  
To better accommodate diverse data characteristics, \textit{personalized federated learning (pFL)} \cite{smith2017federated, arivazhagan2019federated},  \cite{fallah2020personalized} has been developed. 
This approach aims to tailor models to better fit individual client data while still leveraging the collaborative benefits of FL. In pFL, models are partitioned into global and client-specific segments, allowing for both shared learning and individual adaptation. 
For instance, the LG-FedAvg method \cite{liang2020think} begins with the development of a shared global model. Subsequently, only the global portions of the model are aggregated server-side, while clients update their personalized segments locally, thus balancing generalization with personalization.
Another method enhancing pFL, especially in non-IID environments, is FedBN \cite{li2021fedbn}, which incorporates client-specific batch normalization (BN) layers. While most of the model is trained and aggregated globally, these BN layers are tailored locally, adapting the model to each client's unique data distribution. This customization helps manage the distinct statistical properties of diverse datasets, improving both personalization and model effectiveness.

However, the computational and communication constraints of client devices necessitate efficient model designs in pFL.
Inspired by the \textit{lottery ticket hypothesis} \cite{frankle2018lottery}, which suggests that a well-pruned sub-model can match the performance of a full model while being considerably more compact, \textit{model pruning} has been integrated into FL. 
The Hermes algorithm \cite{li2021hermes}, for example, enables clients to prune models locally before aggregation, balancing broad knowledge integration with personalized adjustments. This pruning strategy not only reduces model size but also preserves the customization for individual clients.

Yet, the effectiveness of federated model pruning (FMP) can be limited when client data varies significantly, as traditional FMP methods often assume homogeneity among client data. This assumption does not hold in many practical applications like image recognition or personalized advertising, where data is inherently non-IID. To address these challenges, our paper introduces a novel approach that groups users by data similarity, facilitating the development of pruned models that are both relevant and efficient.

\IEEEpubidadjcol

In this paper, we tackle the issues in FMP by grouping users based on the similarity of their data distributions, which allows for the development of customized models for each cluster, thus improving both the relevance and effectiveness of the pruned models.
Specifically, we propose a two-stage framework to improve the efficiency of model pruning in personalized FL, which is referred to as \textbf{S}tructure-\textbf{A}ware 
\textbf{F}ederated \textbf{L}earning (SAFL). Specifically, the main contributions of this paper are summarized as follows:
\begin{itemize}
    \item SAFL introduces an \textit{iterative clustering} mechanism that groups clients based on the similarity of their data distributions. This clustering is pivotal for identifying clients with similar data characteristics, which allows for more relevant model pruning and training that is tailored to the specific needs of grouped clients.
    
    \item The framework employs a novel strategy named \textit{SCSI (Similar Client Structure Information)-guided model pruning}. In this approach, aggregated pruning criteria from clients within the same cluster guide the pruning process. This method ensures that the pruning is not only data-driven but also cognizant of the structural insights shared among clients in the same cluster, leading to more efficient and effective personalized models.

    \item After SCSI-guided model pruning in the first stage, we introduce the second stage, where the clients train their small-sized pruned models locally with heterogeneous model aggregation. 
    This \textit{two-stage approach} significantly reduces the size of the models deployed on client devices, thus lowering both computation and communication overheads, and allows the integration of personalized models with the global model, ensuring that each client benefits from both personalized and global knowledge. 
    
    \item We conduct experimental validation where SAFL is compared against traditional federated learning methods and other personalized approaches. The results demonstrate that SAFL not only achieves higher accuracy but also significantly reduces model size, thereby reducing computational burden and enhancing performance efficiency.
\end{itemize}

\section{Related Work}
\subsubsection{Model Pruning}
Model pruning has evolved to enhance computational efficiency in federated learning. 
Frankle and Carbin's lottery ticket hypothesis \cite{frankle2018lottery} highlights the discovery of smaller, efficient sub-models (winning tickets) within larger models that maintain original performance, optimizing training in distributed environments. 
Liu et al. critique traditional pruning methods, suggesting that smaller, designed-from-scratch architectures might outperform large, pruned models, shifting pruning from size reduction to architectural optimization for resource-limited settings \cite{liu2018rethinking}. 
He et al. introduce a structured channel pruning via LASSO regression that enhances model adaptability and efficiency \cite{he2017channel}, critical for diverse federated learning scenarios. 
Liu et al.'s \textit{Network Slimming} \cite{liu2017learning} employs L1 regularization on weight parameters in BN layers to induce sparsity, reducing model size while maintaining or improving accuracy, suitable for communication-constrained federated learning. 
These contributions collectively highlight model pruning's pivotal role in refining federated learning models, balancing performance with efficiency in decentralized models.

\subsubsection{Federated Model Pruning}
Pruning techniques in FL have significantly evolved to address the challenges of heterogeneity, model homogeneity, and resource constraints. FedTiny \cite{huang2023distributed} and FedPrune \cite{munir2021fedprune} respectively enhance model efficiency through adaptive BN and progressive pruning, and dynamically adjust the global model for devices with limited capabilities, improving fairness and robustness over non-IID data. 
FedRolex \cite{alam2022fedrolex} reduces model homogeneity by training larger global models and creating personalized sparse sub-models through structured pruning. 
In Hermes \cite{li2021hermes}, each client prunes the model using its local data before the results are aggregated on the server. Hermes employs a novel model aggregation method that aggregates overlapping parameters across the clients' local models while preserving non-overlapping parameters unchanged. 
Similarly, PruneFL \cite{jiang2022model} and FedMP \cite{jiang2022fedmp} introduce adaptive pruning strategies that manage resource constraints effectively, with FedMP employing a multi-armed bandit algorithm for dynamic pruning adjustments and a novel parameter synchronization scheme to ensure convergence.
FL-PQSU \cite{xu2021accelerating} and FedDUAP \cite{zhang2022fedduap} represent advances in integrating structured pruning, weight quantization, and selective updating to optimize neural network models during the FL process, reducing resource demands and improving performance across heterogeneous environments.
These approaches allow each client's local model to incorporate broader knowledge while maintaining personal customizations.

\subsubsection{Clustered Federated Learning}
In federated learning, various frameworks have been developed to address challenges such as data heterogeneity, Byzantine faults, and non-IID data. 
Ghosh et al. \cite{ghosh2019robust} propose a robust FL framework using a novel clustering technique to group models based on data similarity, enhancing learning integrity even in the presence of adversarial nodes. Similarly, Briggs et al. and 
Sattler et al. introduce hierarchical and geometric clustering approaches in their FL frameworks \cite{sattler2020clustered} to efficiently manage non-IID data and adapt to nonconvex objectives without altering the FL communication protocols, respectively.
Further advancing clustered learning, Ghosh et al. present the Iterative Federated Clustering Algorithm (IFCA) \cite{ghosh2020efficient}, which dynamically clusters clients to improve learning efficiency under highly heterogeneous data. 
Liu et al.'s PFA method \cite{liu2021pfa} enhances model personalization by using the sparsity properties of neural networks to create privacy-preserving representations for client grouping.
Wang et al. \cite{wang2021resource} enhance resource efficiency in edge computing by using hierarchical aggregation in federated learning to coordinate local updates within clusters, optimizing communication and privacy.

\begin{figure}[htbp]
\centering
\includegraphics[width=\linewidth]{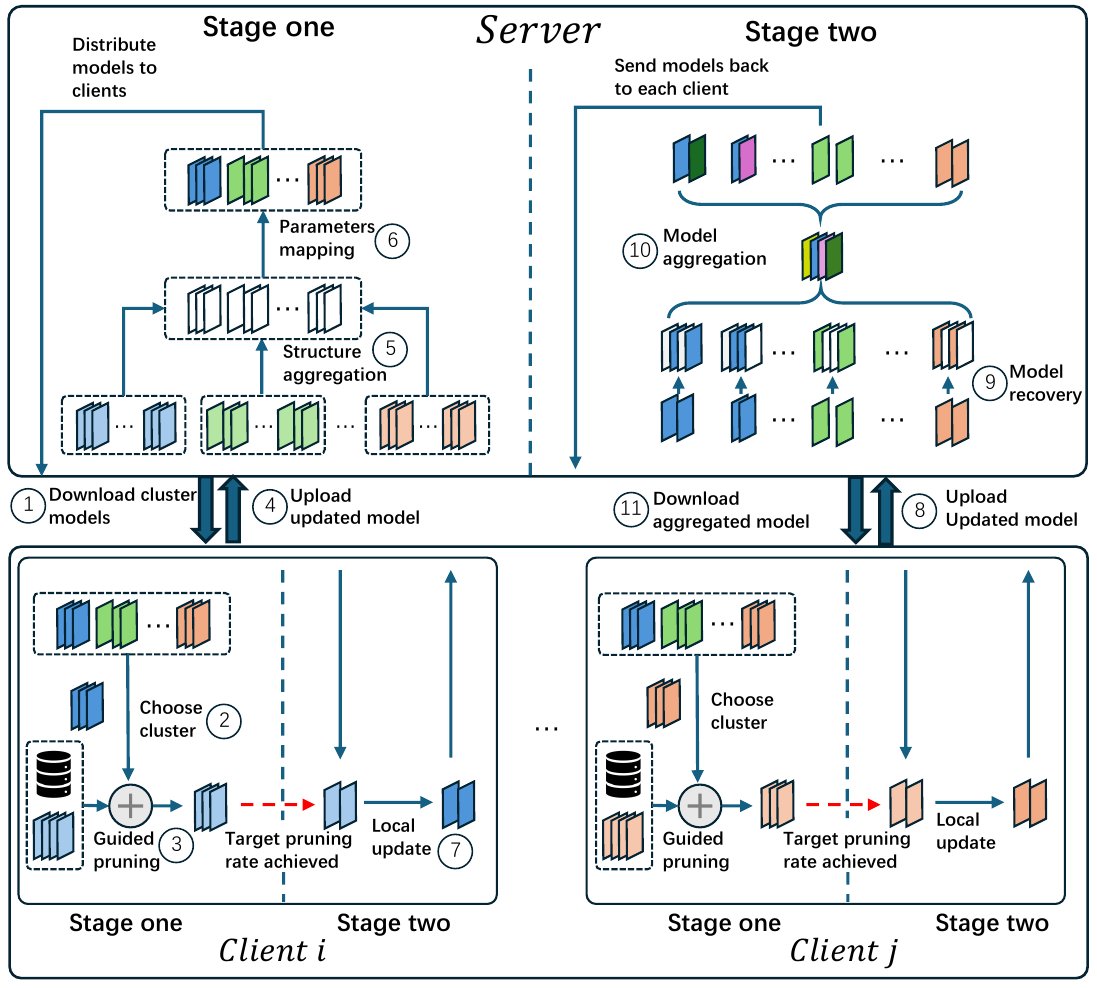} 
\caption{Overview of the SAFL framework.}
\label{fig one}
\vspace{-0.5cm}
\end{figure}

\section{Proposed Method}
\subsection{Overview of the SAFL Framework}

To address the challenge of effective personalization in FMP, where pruned models fail to accurately reflect client task characteristics due to the limited amount of local data and heterogeneous data across clients, we propose the two-state SAFL framework. SAFL clusters clients with similar data and task characteristics, using a novel approach that aggregates their pruning criteria into a cluster model. This model then guides each client in performing informed and effective pruning, ensuring the pruned models better align with the specific requirements of their tasks. 



An illustrative overview of SAFL is depicted in Fig.~\ref{fig one}, the first stage of the process employs an iterative clustering algorithm to divide the clients into $K$ suitable clusters. 
During each iteration, clients perform the following steps: they download $K$ (small-sized) pruned cluster models from the server (\textcircled{\scriptsize{1}}), select the model that best matches their local data requirements (\textcircled{\scriptsize{2}}), and employ SCSI-guided pruning using the structural insights from their chosen model (\textcircled{\scriptsize{3}}). 
Subsequently, the server collects the pruned models from all clients within the same cluster (\textcircled{\scriptsize{4}}). 
After pruning, the server aggregates the models from each cluster $j \in [K]$ using \textit{heterogeneous model fusion}, which {integrates all models within cluster $j$ into a unified cluster model}.
The structure of this consolidated model is determined by the shared channels among the individual models (\textcircled{\scriptsize{5}}), and {the parameters from the client models are averaged and incorporated into the cluster model} (\textcircled{\scriptsize{6}}). 
The new cluster models are then redistributed to the clients for further evaluation and selection in subsequent iterations.
This sequence from (\textcircled{\scriptsize{1}}) to (\textcircled{\scriptsize{6}}) continues until the desired pruning rate is achieved, concluding the first stage with each client obtaining a finely pruned model that closely aligns with their specific task needs.

In the second stage of SAFL, clients further refine the sparse models derived from the first stage by training them with their local data over $\tau$ epochs (\textcircled{\scriptsize{7}}), followed by uploading these refined models to the server (\textcircled{\scriptsize{8}}). The server collects all client models, aligns each client's sub-model to the (full-sized) original model structure (\textcircled{\scriptsize{9}}), and performs personalization-preserving aggregation (\textcircled{\scriptsize{10}}). This process ensures that the aggregated model reflects the unique structural nuances of each client's model, producing heterogeneous models tailored for the clients. These tailored models are then sent back to the clients for subsequent iterations (\textcircled{\scriptsize{11}}). The cycle from (\textcircled{\scriptsize{7}}) to (\textcircled{\scriptsize{11}}) is repeated until the set number of communication rounds is completed.

\begin{algorithm}[t]
\caption{Clustered Model Pruning}\label{alg1}
\SetAlgoNlRelativeSize{-1}
\SetKwInOut{Input}{Input}
\SetKwInOut{Output}{Output}

\Input{total pruning rounds $T\in \mathbb{N}_{+}$; 
number of clusters $K\in \mathbb{N}_{+}$;
pruning rate array $r_0,\dots,r_{T-1} \in [0,1]$; 
initial client model parameters $\bm{\theta}^{\text{client},0}_i$ for $i \in [N]$;  
initial within-cluster model parameters $\bm{\theta}^{\text{cluster},0}_j$ for $j \in [K]$;  
regularization parameters $\lambda,\beta \in \mathbb{R}_{+}$}

\For{$t = 0,1,\dots,T-1$}{
    The server broadcasts $\{\bm{\theta}^{\text{cluster},t}_j\}_{j=1}^{K}$ to all clients; \\
    \For{\textnormal{all client} $i \in [N]$ \textnormal{in parallel}}{
        Estimate cluster identity 
        \[
          \hat{j}^t_i = \arg\min_{j \in [K]} \ell_i(\bm{\theta}^{\text{cluster},t}_j);
        \]\\
        Define one-hot encoding vector $s_i = \{s_{i,j}\}_{j=1}^k$, 
        where $s_{i,j} = 1$ if $j = \hat{j}^t_i$ else $0$;\\
        $\bm{\Theta}^{\text{client},t}_i = \mathsf{ModelRecover}({\bm{\theta}}^{\text{client},t}_i)$\\
        $\widetilde{\bm{\Theta}}^{\text{client},t}_i = \mathsf{GuidedUpdate}\bigl(\bm{\Theta}^{\text{client},t}_i,\,
        \bm{\theta}^{\text{cluster},t}_{\hat{j}^t_i}\bigr)$\\
        ${\bm{\theta}}_i^{\text{client},t} = \mathsf{NetSlim}(\widetilde{\bm{\Theta}}_i^{\text{client},t},r_t)$\\
        ${\bm{\theta}}_i^{\text{client},t} = \mathsf{FineTune}\bigl({\bm{\theta}}_i^{\text{client},t},\tau\bigr)$\\
        send ${\bm{\theta}}_i^{\text{client},t}$ to the server;
    }
    \textit{Structure Aggregation:} The server aggregates the structures of 
    clients' models within each cluster $j$ into $\bm{\theta}_j^{\text{cluster},t+1}$ for all $j \in [K]$.\\
    \textit{Parameter Mapping:} The server then maps the parameters of 
    clients' models within each cluster $j$ according to:
    \[
        \bm{\theta}_j^{\text{cluster},t+1} = \sum_{i=1}^{N} \frac{s_{i,j} {\bm{\theta}}_i^{\text{client},t}}{\sum_{i=1}^{N} s_{i,j}}
        \quad \text{for all} \quad j \in [K];
    \]
}

\Output{${\bm{\theta}}_i^{\text{client},T-1}$ for all $i \in [N]$}

\end{algorithm}

\subsection{Stage One: Clustered Model Pruning
}

The primary objective of SAFL's initial stage is to address the issue of insufficient local data in pruned federated learning. It achieves this by clustering similar clients, ensuring that each sub-model accurately reflects its specific task characteristics.
Details of the algorithm in stage one are presented in Algorithm \ref{alg1}, which consists of the following subroutines.

\subsubsection{Client Clustering}
The client clustering of SAFL, as presented in Lines 4--5 of Algorithm \ref{alg1}, is based on the iterative method IFCA \cite{ghosh2020efficient}. 
In each round $t$, each client receives models $\bm{\theta}^{\text{cluster},t}$ from all $K$ clusters and determines its cluster identity by identifying the model that yields the lowest loss:
\begin{align}
    \hat{j}^t_i = \arg\min_{j \in [K]} \ell_i(\bm{\theta}^{\text{cluster},t}_j),
    \nonumber
\end{align}
where $\ell_i$ is the loss function of client $i$ defined for the pruned cluster model $\bm{\theta}^{\text{cluster},t}_j$ and $\hat{j}^t_i$ is the cluster index of client $i$ in round $t$.
We remark that the cluster identities of the clients are iteratively refined during Algorithm \ref{alg1}, which ensures continuous adaptation of the cluster models to suit the needs of each client.


\begin{figure}[t]
\centering
\includegraphics[width=\linewidth]{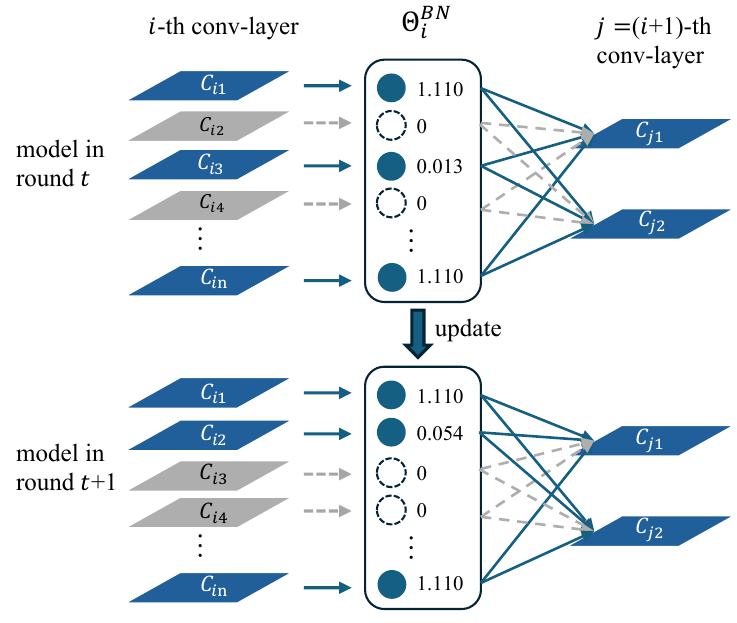} 
\caption{Model recovery technique in stage one.}
\label{fig two}
\vspace{-0.5cm}
\end{figure}

\subsubsection{Model Recovery}
In iterative model pruning, the significance of channels can shift dynamically throughout the pruning process. 
This variability can result in scenarios where channels initially deemed less important gain significance in later stages. 
As a result, prematurely pruning these channels can adversely affect the model's overall performance.

To address this, we have introduced a model recovery approach within our SAFL framework, implemented through the $\mathsf{ModelRecover}$ subroutine outlined in Algorithm \ref{alg1}.
Fig.~\ref{fig two} shows an example of $\mathsf{ModelRecover}$ being applied to convolutional neural networks across two consecutive rounds. 
Before each session of local training, the pruned model of a client is restored to its full size. 
The weights in the BN layer, denoted by $\bm{\Theta}^{\text{BN}}_i$, of the pruned channels (in grey color) are initially set to be zero.
This process allows for the correction of previously mispruned channels; for instance, a channel $C_{i2}$ that was pruned in round $t$ might be updated and retained in round $t+1$.

\subsubsection{Guided Model Update}

As demonstrated by the FedBN method \cite{jiang2022fedmp} in pFL, the structure of a pruned model is significantly influenced by the BN layer, represented by $\bm{\Theta}^{\text{BN}}_i$ for $i \in [N]$. 
Drawing inspiration from this, in our SAFL framework, we cluster clients with similar data distributions. This clustering enables clients to share and leverage structural information encapsulated in $\bm{\Theta}^{\text{BN}}_i$, facilitating the identification of suitable sub-model structures even with limited local data.

Within these clusters, clients exchange insights and learn from the structural nuances of each other's models. 
To enhance this process, we introduce a regularization term in SAFL specifically designed to minimize the discrepancies in the $\bm{\Theta}^\text{BN}_i$ values across models within the same cluster. 
After determining their cluster affiliations, clients proceed to train their models using their local datasets.
Specifically, we first apply the $\mathsf{ModelRecover}$ subroutine to obtain the $\hat{j}^t_i$-th full-sized cluster model:
\[
    \bm{\Theta}^{\text{cluster},t}_{\hat{j}^t_i} = \mathsf{ModelRecover}(\bm{\theta}^{\text{cluster},t}_{\hat{j}^t_i}).
\]
The BN layer of this reconstructed model, specifically the $b$-th BN layer, is represented by $\bm{\Theta}^{\text{clusterBN},t}_{\hat{j}^t_i,b}$.

Using this reconstructed cluster model and the guiding cluster model $\bm{\theta}^{\text{cluster},t}_{\hat{j}^t_i}$, the $\mathsf{GuidedUpdate}$ subroutine is employed as specified in Line 7 of Algorithm \ref{alg1}. The training of the new client model $\bm{\Theta}_i^{\text{client},t}$ for each client $i$ involves a loss function structured as follows:
\begin{align}
    &\mathcal{L}_i(\bm{\Theta}_i) + \lambda \sum_{b=1}^{B} \| \bm{\Theta}^{\text{BN}}_{i,b} \|_1 +
    \mu \sum_{b=1}^{B} \left\| \bm{\Theta}^{\text{BN}}_{i,b} - \bm{\Theta}^{\text{clusterBN},t}_{\hat{j}^t_i,b} \right\|_1,
\nonumber
\end{align}
where $\mathcal{L}_i(\bm{\Theta}_i) \coloneqq \frac{1}{M_i} \sum_{m=1}^{M_i} \mathcal{L}_i(\bm{\Theta}_i; \bm{z}_i)$ is the loss function for client $i$, defined over the local dataset $\{\bm{z}\}_{m=1}^{M_i}$ for the full-sized model $\bm{\Theta}_i$, $B$ is the number of BN layers, and $\mu$ is the regularization parameter.
After each client's distributed training according to this loss function, the subroutine $\mathsf{GuidedUpdate}$ outputs the full-sized client local model, denoted by $\widetilde{\bm{\Theta}}^{\text{client},t}_i$.
This loss function aims to tailor each client's model parameters to their local data while aligning the BN layers closely with those of the cluster model. The cluster model $\bm{\Theta}^{\text{clusterBN},t}_{\hat{j}^t_i,b}$, which aggregates all client models from the previous round within the cluster, ensures collective contributions significantly shape the evolution of the cluster model.

\subsubsection{Model Pruning}
After each client's model is recovered and updated, it is subjected to another round of pruning to meet the targeted pruning rate for that iteration.
This approach ensures that potentially valuable channels are preserved, thereby improving the model's adaptability and accuracy through successive pruning cycles. Pruning can often lead to a significant decrease in model accuracy, which adversely affects subsequent model aggregation and, ultimately, the overall performance of the cluster model. Therefore, multiple rounds of fine-tuning are essential after each pruning phase to restore and enhance the model's performance and accuracy.

Pruning algorithms in neural networks fall into two categories: unstructured and structured. Unstructured pruning \cite{han2015deep}, \cite{han2016eie}, \cite{lee2018snip} removes individual weights, enhancing performance flexibility but increasing computational demands. Structured pruning \cite{wen2016learning}, \cite{you2019gate}, \cite{nonnenmacher2021sosp}, on the other hand, eliminates entire channels or filters, streamlining the model and reducing computational needs, which is ideal for environments with limited resources but offers less flexibility.
The model pruning subroutine in SAFL, $\mathsf{NetSlim}$ as outlined in Line 8 of Algorithm \ref{alg1}, is built upon the \textit{Network Slimming} method \cite{liu2017learning}, a well-established method of structured pruning.
In the $\mathsf{NetSlim}$, client ranks the weights in the $\widetilde{\bm{\Theta}}^{\text{BN}, t}_i$. The threshold is set at the $r_t$-th smallest weight.
Channels whose weights exceed this threshold are retained, while those falling below are pruned.
After each client's distributed training according to this loss function, the subroutine $\mathsf{NetSlim}$ outputs the small-sized pruned client model, denoted by $\bm{\theta}^{\text{client},t}_i$.

As the pruned models $\bm{\theta}^{\text{client},t}_i$ might suffer from reduced generalization performance due to their diminished size, these models are further fine-tuned for $\tau$ epochs. This additional fine-tuning is carried out to ensure enhanced performance of the pruned models and is conducted according to the $\mathsf{FineTuning}$ subroutine detailed in Algorithm \ref{alg1}.

\subsubsection{Heterogeneous Model Fusion}
Each client sends their fine-tuned model to the server, where the server aggregates all the heterogeneous models within a cluster. To effectively combine heterogeneous pruned clients' models within the same cluster and obtain a unified cluster model with an identical number of channels to guide client pruning, we have developed a heterogeneous model fusion method.

\begin{figure}[t]
\centering
\includegraphics[width=0.95\linewidth]{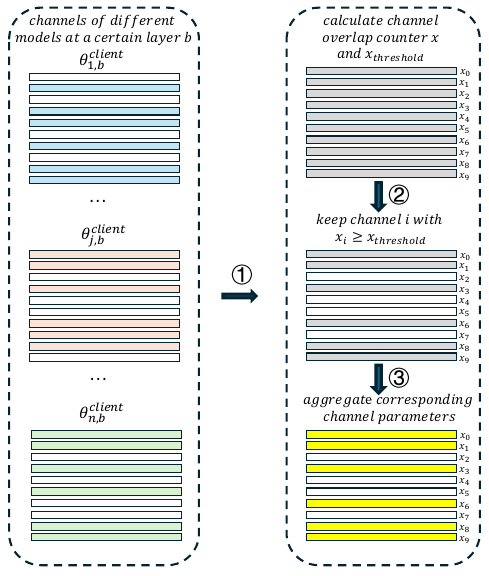} 
\caption{Heterogeneous model fusion in stage one.}
\label{fig three}
\vspace{-0.3cm}
\end{figure}

As illustrated in Fig.~\ref{fig three}, we compute the \textit{overlapping frequency} for each channel within every layer across all models in the cluster to obtain the overlap counter $x$ (\textcircled{\scriptsize{1}}). This frequency measures how often the same channel appears across different models. Based on the target pruning rate for the round, we establish a threshold $x_{threshold}$ and preserve channels $i$ where $x_i > x_{threshold}$ (\textcircled{\scriptsize{2}}). The parameters of these selected channels are then determined by averaging the corresponding parameters from all models in the cluster that include these channels (\textcircled{\scriptsize{3}}). This aggregated cluster model subsequently serves as one of the initial models for the clusters, which each client will select in the next iteration.


\subsection{Stage Two: Personalization-Preserving Aggregation}
\label{subsect: aggregation}

Once the target pruning rate is achieved and the cluster partition has stabilized, SAFL proceeds with stage two. We employ the personalization-preserving aggregation method proposed in Hermes \cite{li2021hermes}, where each client trains its heterogeneous model while the server executes heterogeneous model aggregation, as described in Algorithm \ref{alg2}.

\begin{figure*}[t]
\centering
\includegraphics[width=0.98\textwidth]{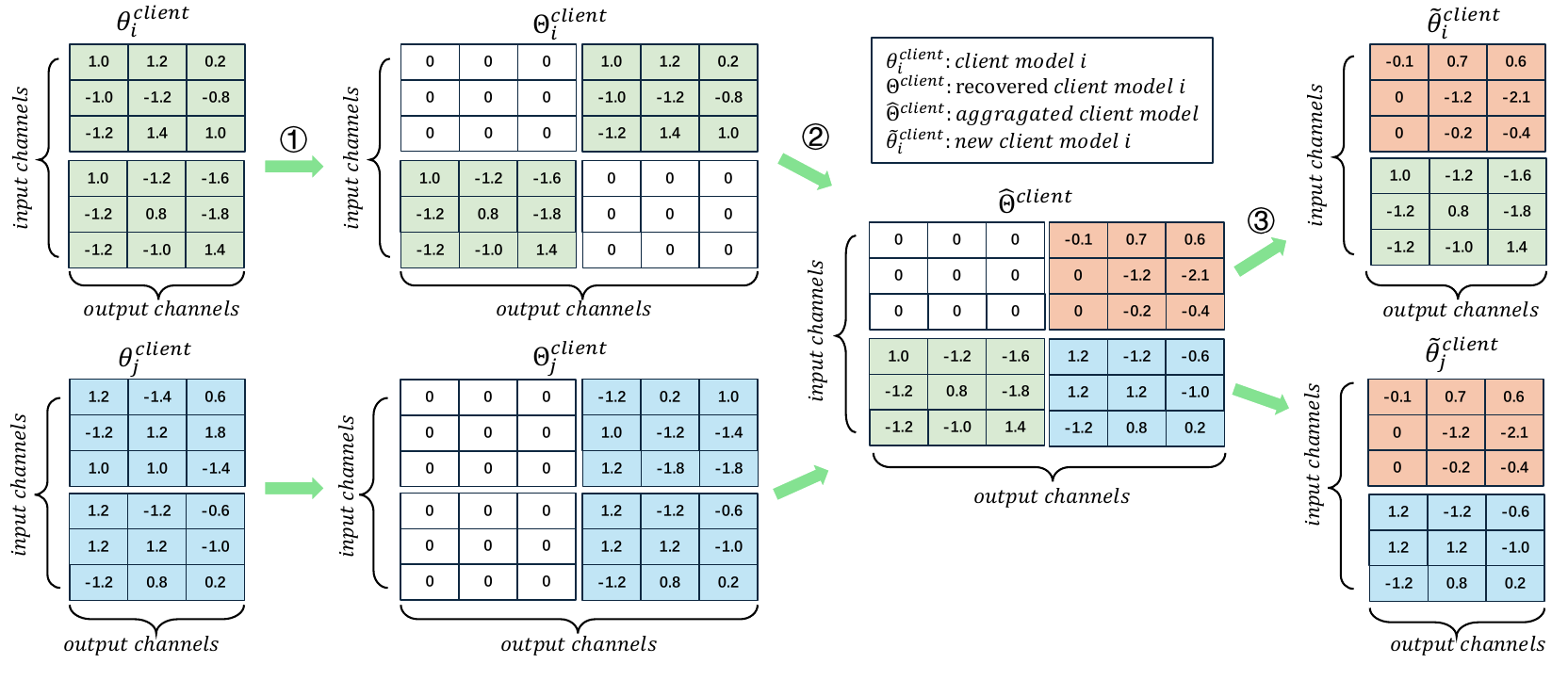} 
\caption{Procedures of heterogeneous model aggregation in stage two.}
\label{fig four}
\vspace{-0.3cm}
\end{figure*}

As shown in Fig.~\ref{fig four}, each client locally updates its model using $\tau$ iterations of Stochastic Gradient Descent (SGD) and then transmits the updated model to the server for aggregation. Initially, the server restores each client’s model to its full size (\textcircled{\scriptsize{1}}), initializing the parameters of the reinstated channels to zero. Because the structure of the clients' models does not change during this phase, the pruning masks are retained on the server to reduce communication overhead. Subsequently, the server aggregates the models using techniques similar to FedAvg (\textcircled{\scriptsize{2}}). After completing the aggregation, the server reconstructs each client’s model 
and sends it back to the client (\textcircled{\scriptsize{3}}). This process ensures that each client’s model retains robust generalization capabilities through the shared parameters, while its unique features are preserved via the distinct, non-overlapping parameters. Moreover, since the model size transmitted is significantly smaller than the original, this method substantially reduces communication overhead.


\begin{algorithm}[t]
\caption{Aggregating Heterogeneous Sub-Models}\label{alg2}
\SetAlgoNlRelativeSize{-1}
\SetKwInOut{Input}{Input}
\SetKwInOut{Output}{Output}

\Input{Number of global iteration rounds $G$}
\Output{Final client models $\bar{\bm{\theta}}_i^{\text{client},G-1}$ for each client $i \in [N]$}

\textbf{Initialize:} Client $i$'s model 
$\bar{\bm{\theta}}_i^{\text{client},0} = 
{\bm{\theta}}_i^{\text{client},T-1}$ 
for $i \in [N]$
\\
\For{$t = 0,1,\dots,G-1$}{
    \For{\textnormal{all client} $i \in [N]$ \textnormal{in parallel}}{
        $\bar{\bm{\theta}}^{\text{client},t}_i 
         = \mathsf{LocalTraining}\bigl(\bar{\bm{\theta}}^{\text{client},t}_i,\eta\bigr)$
    }
    The server performs the personalization-preserving aggregation 
    mentioned in~\ref{subsect: aggregation}.
}
\end{algorithm}

\subsection{Analysis of Communication Overhead}
The SAFL framework introduces mechanisms to effectively reduce communication overhead, which is critical in federated learning environments where network resources are often limited. This section quantitatively analyzes these mechanisms and presents empirical results that highlight the efficiency of SAFL in managing communication costs.

Assuming $N$ clients and $K$ clusters, the original model $\bm{\Theta}$ has a size of $M$. The pruning rate array $r$ contains $T$ elements, corresponding to the number of rounds in stage one of the SAFL framework, while stage two comprises $G$ rounds. Assume the size of the pruned model with pruning rate $r_i$ is $(1 - r_i) M$, and let $p_i = (1-r_i)$.

\subsubsection{Initial Model Distribution}
Initially, the server distributes the original model $\bm{\Theta}^{\text{client}}$ and $K$ cluster models $\bm{\Theta}^{\text{cluster}}$ to each client. Assuming that each client receives a complete copy of the model, the total communication cost for the initial distribution is:
\[
    C_{\text{distribution}} = N M + K N M = (K + 1) N M.
\]

\subsubsection{Clustered Model Pruning}
In each round of stage one, each client receives $K$ cluster models and then sends their pruned model to the server.
During the first iteration, clients receive cluster models of the original size $M$.
In subsequent iterations $i$, clients receive cluster models with a reduced size of $p_{i-1} M$.
\begin{align*}
    C_{\text{download}} &= N K M + \sum_{i=0}^{T-2} p_i N K M = N K M (1 + \sum_{i=0}^{T-2} p_i),
    \\
    C_{\text{upload}} &= \sum_{i=0}^{T-1} p_i N M = N M \sum_{i=0}^{T-1} p_i.
\end{align*}

\subsubsection{Personalization-Preserving Aggregation}
During this phase, each client performs local updates on its personalized pruned model for \(G\) rounds. After each round, the client uploads its updated model to the server for aggregation and subsequently downloads the aggregated model. The communication overhead incurred during this process is given by:
\[
    C_{\text{update}} = 2 G M N p_{T-1}.
\]
Therefore, the total communication cost of SAFL is:
\begin{align*}
    C_{\text{SAFL}} =&~ C_{\text{distribution}} + C_{\text{download}} + C_{\text{upload}} + C_{\text{update}} \\
    =&~ (K + 1) N M + N K M (1 + \sum_{i=0}^{T-2} p_i) \\
    & + N M \sum_{i=0}^{T-1} p_i + 2 G M N p_{T-1}\\
    =&~ N M \left(2K + 1 + K \sum_{i=0}^{T-2} p_i + \sum_{i=0}^{T-1} p_i + 2 G p_{T-1}\right).
\end{align*}

\subsubsection{Comparison with FedAvg}
In FedAvg, each client uploads and downloads the model for $G$ rounds. Assuming no pruning and thus the full model size $M$ is used, the communication overhead for FedAvg is:
\[
    C_{\text{FedAvg}} = 2 G M N.
\]

Assuming both Stage Two of SAFL and FedAvg involve $G$ communication rounds, to determine when SAFL is more efficient, we set:

\[
N M \left(2K + 1 + K \sum_{i=0}^{T-2} p_i + \sum_{i=0}^{T-1} p_i + 2 G p_T\right) < 2 G M N.
\]
Solving for \(G\) gives
\[
G > \frac{2K + 1 + (K+1) \sum_{i=0}^{T-2} p_i + p_{T-1}}{2 (1 - p_T)}.
\]

Given that the pruning rates $p_0, \dots, p_{T-1}$ are bounded within $[0,1]$, and both the number of rounds $T$ and the number of clusters $K$ are reasonably small, the typical magnitude of $G$—for example, FedAvg often requires over a thousand rounds to converge on the MNIST dataset under a Non-IID scenario with a small CNN—ensures that the condition for $G$ is readily satisfied. Thus, it can be concluded that SAFL significantly enhances communication efficiency in federated learning environments.

\subsection{Discussion}
\textbf{Different Pruning Methods.} In this work, we employ Network Slimming, which uses the magnitudes of the $\bm{\Theta}^{\text{BN}}$ parameters as pruning criteria. $\bm{\Theta}^{\text{BN}}_i$ is an \(N\)-dimensional vector, representing the number of channels in a layer and functioning as a one-dimensional array for computation. This simplifies alignment and accelerates computations during model training. For instance, computing the L1-norm between $\bm{\Theta}^{\text{BN}, t}_{i,b}$ and $\bm{\Theta}^{\text{clusterBN},t}_{\hat{j}^t_i,b}$ is efficient with a time complexity of \(O(N)\). In contrast, weight pruning evaluates the magnitude of two-dimensional matrices from convolutional kernels, indexed by \(N \times M\), increasing computational complexity to \(O(N \times M)\) and the computational burden on clients. Therefore, Network Slimming is the preferred method in SAFL for balancing computational overhead and model performance.


\textbf{Selection of cluster number \(K\).} SAFL outperforms baselines across various \(K\) values, shown in Tables~\ref{tab:table1} and \ref{tab:table2} in Section~\ref{evaluation}, yet the optimal choice of \(K\) is essential for peak performance. Typically selected based on prior knowledge of the clients' data distributions, \(K\) can be adjusted by initially running SAFL with various \(K\) values, refining it based on clustering outcomes after the cluster structure stabilizes.

\textbf{Different pruning rates across clients.}
SAFL recognizes the diverse computational capabilities of clients in a network, allowing each to adopt a pruning rate suited to its computational power, thereby optimizing performance while accommodating individual resource constraints.

\section{Evaluation}
\label{evaluation}
In this section, we present the experimental setup and numerical results of SAFL and its comparative algorithms, highlighting outcomes with varying numbers of clusters and different pruning rates. 

\subsection{Baselines}
We compare SALF against four baseline methodologies:
\begin{enumerate}
    \item \textbf{FedAvg} \cite{mcmahan2017communication}: 
    A cornerstone federated learning algorithm where each client performs updates on its local model, sends this updated model to a central server for aggregation, and retrieves the aggregated model to continue further updates until convergence is achieved.
    \item \textbf{LG-FedAvg} \cite{liang2020think}: 
    An adaptation of the traditional FedAvg that aggregates model updates on a layer-wise basis instead of aggregating the entire model simultaneously. This method allows for differential updating of model layers, reflecting the distinct contributions of each participating client.
    \item \textbf{FedBN} \cite{li2021fedbn}:
    Modifies the conventional federated learning framework by aggregating model weights across clients while permitting each client to keep its own BN parameters. This strategy is designed to address data heterogeneity by tailoring normalization to the specific data distribution of each client.
    \item \textbf{Hermes} \cite{li2021hermes}: 
    Advances federated learning by allowing clients to maintain personalized local models while contributing to a collective global model. This dual-update approach effectively balances personalization with generalization, integrating both local and global updates.
\end{enumerate}

\subsection{Dataset and Models}
We evaluate the performance of the compared algorithms using the MNIST and CIFAR-10 datasets for both training and testing purposes.

\begin{figure}[t]
\centering
\includegraphics[width=0.95\linewidth]{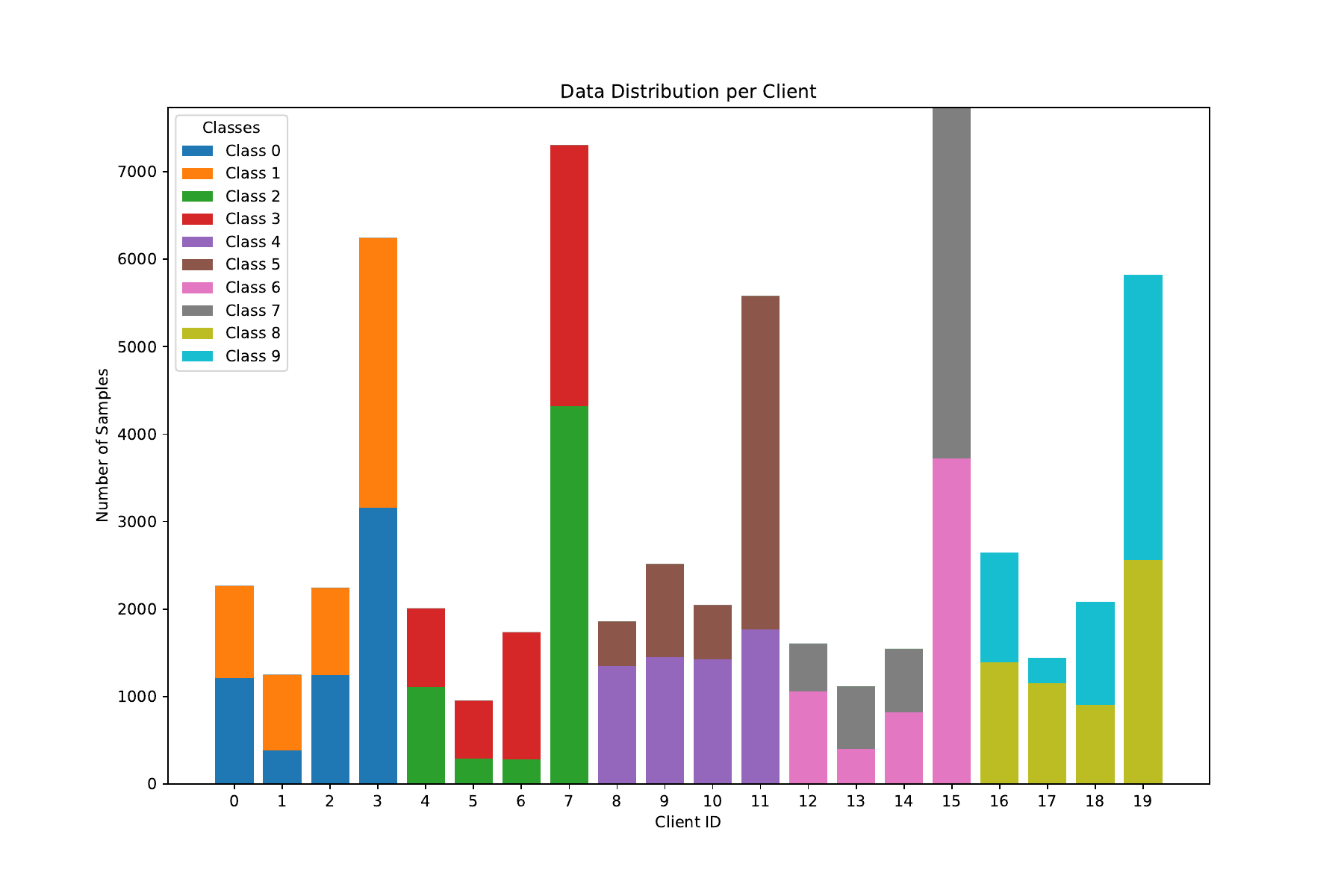} 
\caption{Data partition across clients on Cifar-10 dataset.}
\label{fig five}
\vspace{-0.5cm}
\end{figure}

For the CIFAR-10 dataset, each client possesses data from two specific classes, with the actual classes differing among clients. Additionally, the quantity of data within each class varies, as presented in Fig. \ref{fig five}, leading to an unbalanced distribution on each client. The test data mirrors the distribution found in the training data. The model used for training on CIFAR-10 is a convolutional neural network comprising four convolutional layers. Each convolutional layer is followed by a BN layer, a ReLU activation, and a max pooling layer, concluding with a fully connected layer at the end.

\begin{figure}[t]
\centering
\includegraphics[width=0.95\linewidth]{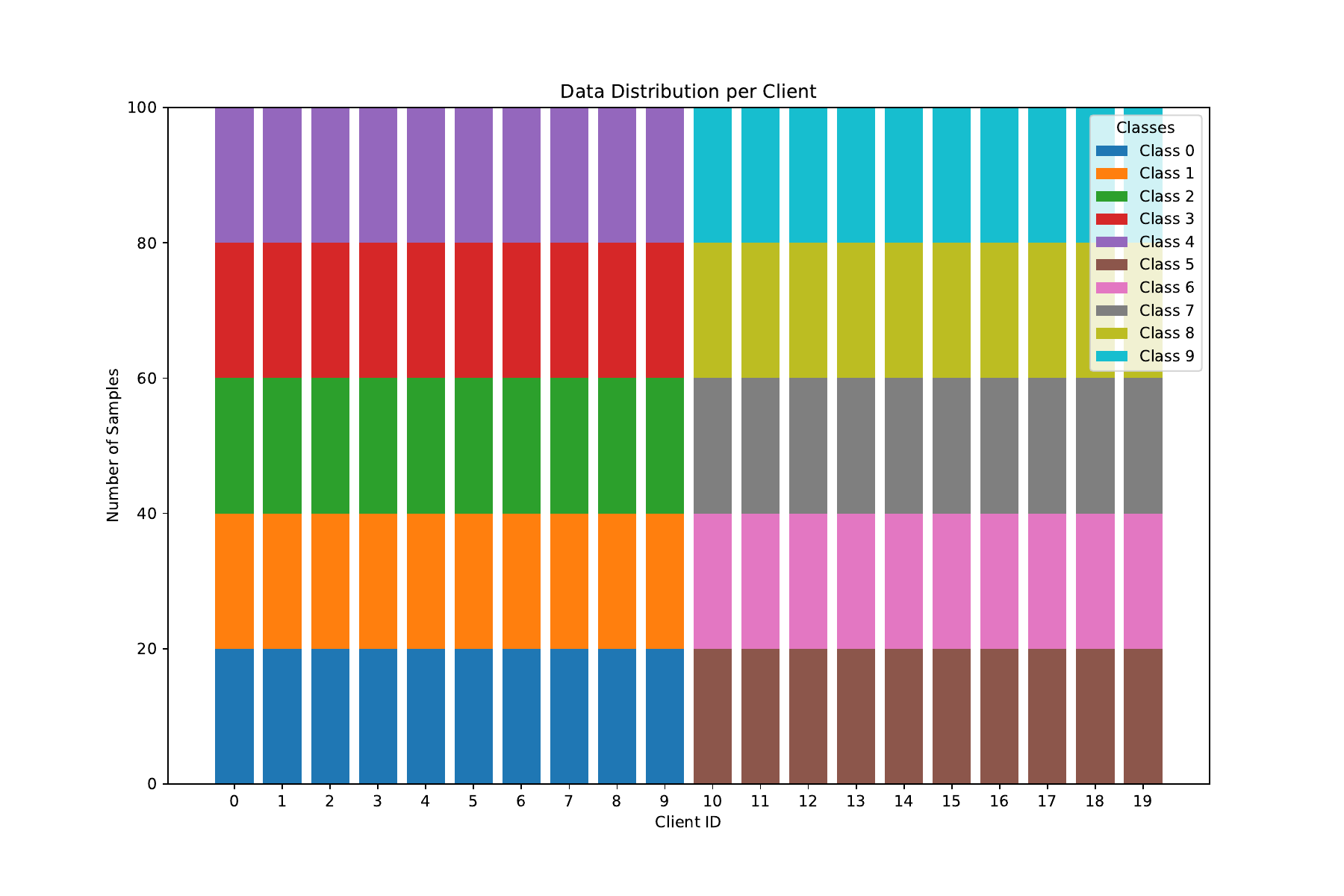} 
\caption{Data partition across clients on MNIST-10 dataset.}
\label{fig six}
\vspace{-0.5cm}
\end{figure}

For the MNIST dataset, as shown in Fig. \ref{fig six}, each client holds data from five distinct classes, with each class containing 20 images. The distribution of the test data mirrors that of the training data. The model utilized for MNIST training includes two convolutional layers. Following each convolutional layer, there is a BN layer, a ReLU activation layer, and a max pooling layer. The final layer of the model is a fully connected layer.

\subsection{Experimental Setup}

In this study, we ensure consistency in the base model configuration and training settings across all baseline methods, including SAFL. We utilize SGD as the optimizer with a learning rate of $0.005$, and employ cross-entropy as the loss function during local training phases, setting the number of epochs to $1$.
Specifically, for both Hermes and SAFL, we implement Network Slimming as the model pruning methodology. The channel sparsity regularization parameter $\lambda$ is set at $0.0001$. The training process dedicated to updating the BN parameters, $\bm{\Theta}^{\text{BN}}_i$, spans $50$ epochs, which is then followed by $20$ epochs of fine-tuning. The pruning rates are incrementally increased in steps of $0.1$; for instance, to achieve a pruning rate of $0.5$, the pruning rate sequence, $r_t$, would be $[0, 0.1, 0.2, 0.3, 0.4, 0.5]$.
In the SAFL framework, the regularization parameter $\mu$, which is used to minimize the divergence between the client model and the cluster model, is set at $0.004$. The experiments are conducted over $1000$ rounds on the CIFAR-10 dataset and $2000$ rounds on the MNIST dataset.

Our findings indicate that aggregating parameters from BN layers leads to significant accuracy reductions when processing non-IID data. Consequently, in line with the approach utilized in FedBN, updates to the BN layers are managed locally at each client without necessitating server communication or aggregation. Given that FedBN is an adaptation of FedAvg that incorporates this strategy, we omit the BN layers from the model in our implementation of the FedAvg algorithm.

\begin{figure*}[t]
\centering
\begin{minipage}{0.46\textwidth}
    \includegraphics[width=0.98\linewidth]{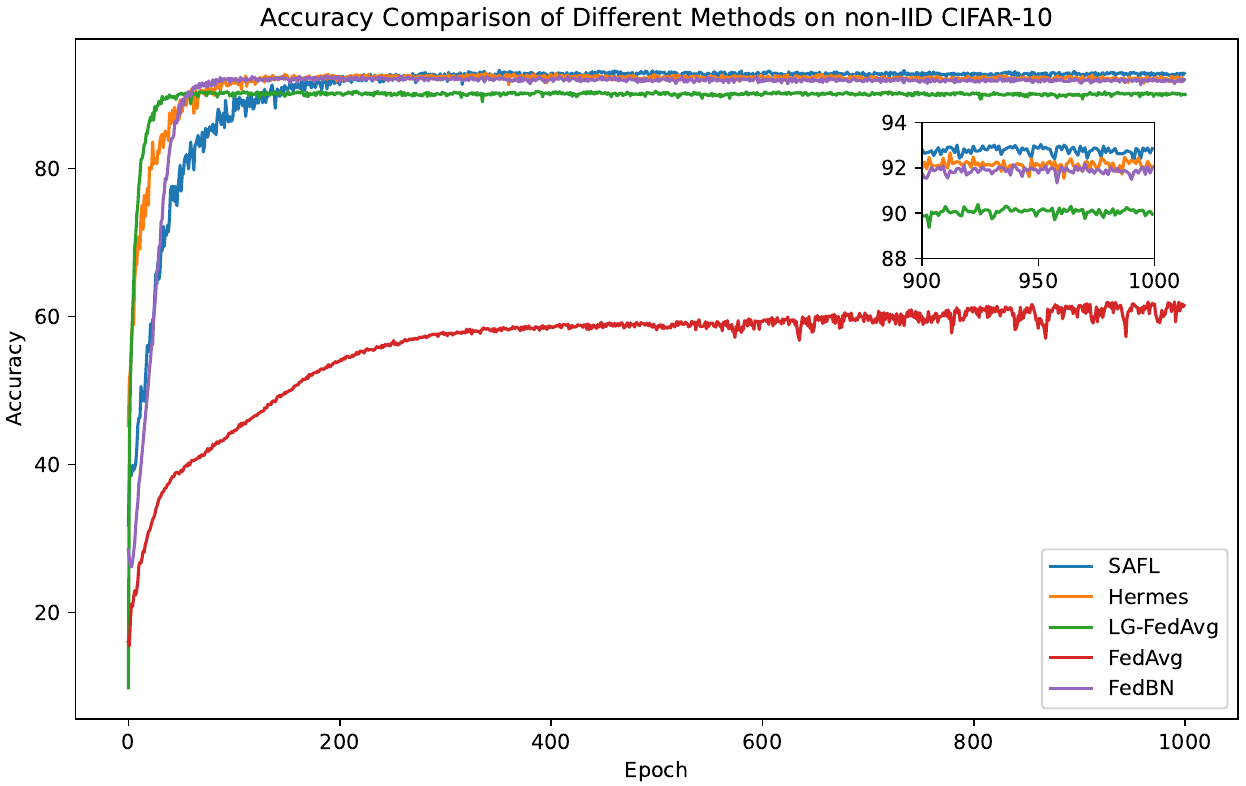}
\end{minipage}
\begin{minipage}{0.46\textwidth}
    \includegraphics[width=0.98\linewidth]{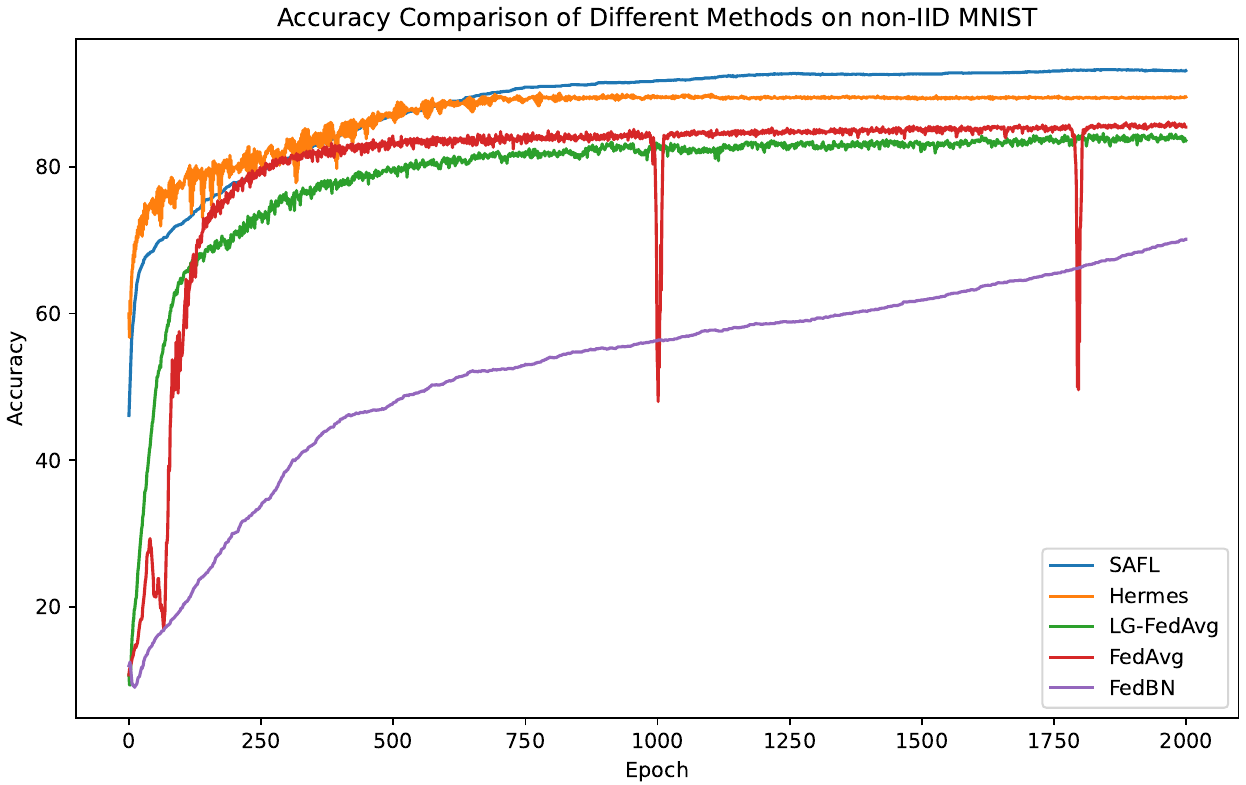}
\end{minipage}
\caption{Training curves on Cifar-10 (left) and MNIST (right) datasets.}
\label{fig seven}
\vspace{-0.2cm}
\end{figure*}

\begin{table}[t]
\fontsize{8}{11}\selectfont
\setlength\tabcolsep{20pt}
\caption{Test accuracies on Cifar-10 dataset}
\centering
\begin{tabular}{l|c}
\toprule
\textbf{Algorithm} & {\textbf{Accuracy} (\%) $\pm$ \textbf{std}} \\
\midrule
FedAvg & 61.98 $\pm$ 0.11 \\
LG-FedAvg & 90.56 $\pm$ 0.11 \\
FedBN & 92.61 $\pm$ 0.11 \\
Hermes ($pr=0.3$) & 92.73 $\pm$ 0.11 \\
SAFL ($pr=0.3, k=1$) & 93.02 $\pm$ 0.31 \\
SAFL ($pr=0.3, k=2$) & 93.04 $\pm$ 0.14 \\
SAFL ($pr=0.3, k=3$) & 93.16 $\pm$ 0.17 \\
SAFL ($pr=0.3, k=4$) & 93.26 $\pm$ 0.18 \\
\textbf{SAFL ($pr=0.3, k=5$)} & \textbf{93.29 $\pm$ 0.13} \\
\bottomrule
\end{tabular}
\label{tab:table1}
\end{table}

\begin{table}[t]
\fontsize{8}{11}\selectfont
\setlength\tabcolsep{20pt}
\caption{Test accuracies on MNIST dataset}
\centering
\begin{tabular}{l|c}
\toprule
\textbf{Algorithm} & \textbf{Accuracy} (\%) $\pm$ \textbf{std} \\
\midrule
FedAvg & 86.02 $\pm$ 0.09 \\
LG-FedAvg & 84.39 $\pm$ 0.16 \\
FedBN & 70.13 $\pm$ 0.05 \\
Hermes ($pr=0.3$) & 90.65 $\pm$ 1.04 \\
SAFL ($pr=0.3, k=1$) & 93.09 $\pm$ 0.02 \\
\textbf{SAFL ($pr=0.3, k=2$)} & \textbf{93.29 $\pm$ 0.02} \\
\bottomrule
\end{tabular}
\label{tab:table2}
\vspace{-0.3cm}
\end{table}

\begin{figure*}[t]
\centering
\includegraphics[width=0.92\textwidth]{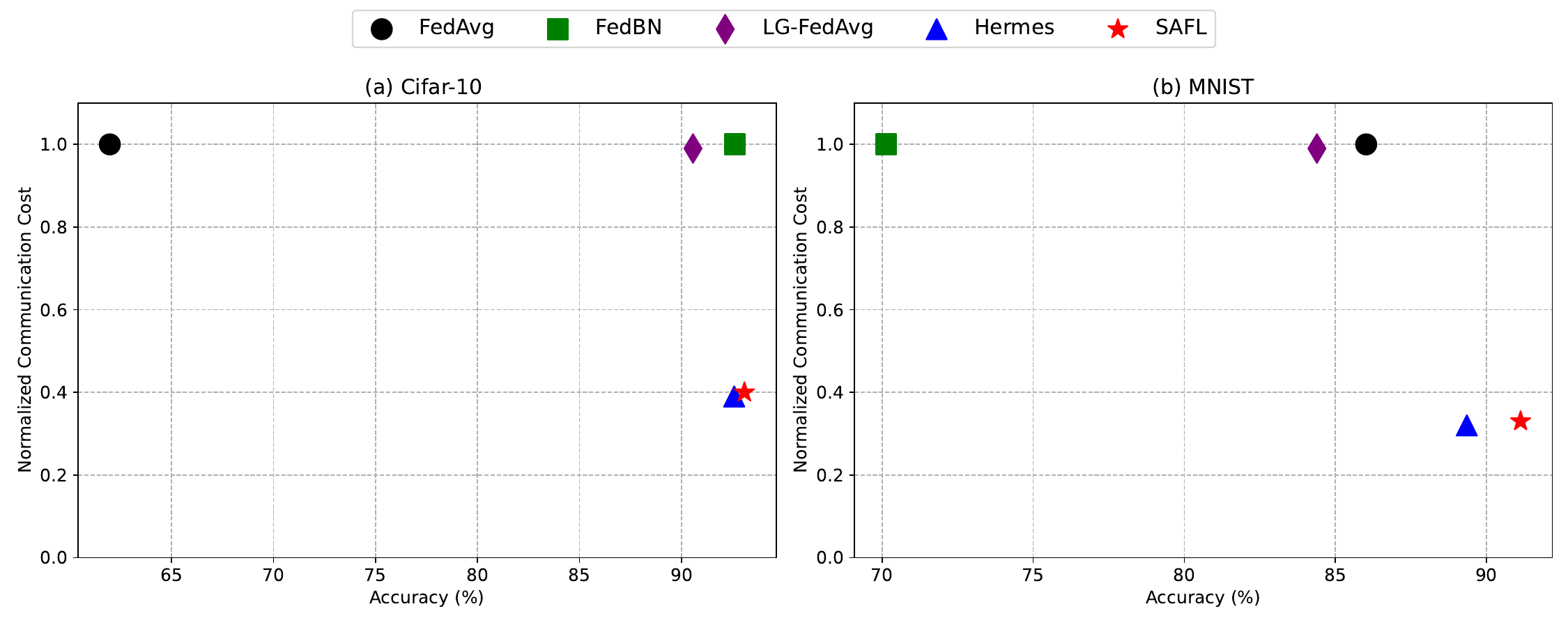}
\caption{Comparison between SAFL and baselines in inference accuracy-communication cost space.}
\label{fig eight}
\vspace{-0.3cm}
\end{figure*}

\begin{table}[t]
\fontsize{8}{11}\selectfont
\setlength\tabcolsep{10pt}
\caption{Comparison between SAFL and Hermes with different pruning rates}
\centering
\begin{tabular}{ccc|cc}
\toprule
\textbf{Dataset} & \makecell{\textbf{Pruning} \\ \textbf{Rate}} & \makecell{\textbf{Model} \\ \textbf{Size}} & \makecell{\textbf{SAFL} \\ \textbf{Accuracy}} & \makecell{\textbf{Hermes} \\ \textbf{Accuracy}} \\
\midrule
\multirow{5}{*}{CIFAR-10} 
& 0.3 & 2.6MB & 93.29\% & 92.73\% \\
& 0.4 & 1.8MB & 93.08\% & 92.58\% \\
& 0.5 & 1.2MB & 92.32\% & 91.67\% \\
& 0.6 & 601KB & 91.13\% & 82.12\% \\
& 0.7 & 289KB & 89.85\% & 80.64\% \\
\midrule
\multirow{4}{*}{MNIST} 
& 0.3 & 509KB & 93.29\% & 90.65\% \\
& 0.4 & 384KB & 92.45\% & 90.25\% \\
& 0.6 & 164KB & 92.24\% & 89.35\% \\
& 0.65 & 101KB & 91.54\% & 86.78\% \\
\bottomrule
\end{tabular}
\label{tab:table3}
\vspace{-0.4cm}
\end{table}


\subsection{Numerical Results on Inference Accuracies}
Inference accuracy is assessed on each client's test data, with results averaged across all clients. For evaluations, pFL methods are evaluated using clients' personalized models, whereas FedAvg is assessed using a unified global model.
From Table~\ref{tab:table1}, Table~\ref{tab:table2} and Figure~\ref{fig seven}, it is evident that SAFL surpasses traditional baselines. 
Compared to FedAvg, LG-FedAvg, and FedBN, SAFL not only achieves higher inference accuracy but also utilizes a model that is 30\% smaller. 
Specifically, on a non-IID CIFAR-10 dataset, SAFL shows an increase in inference accuracy by 20.12\%, 2.73\%, and 0.68\%, respectively, with a pruning rate set at 30\% and five clusters. 
For the non-IID MNIST dataset, the accuracy improvements are 7.27\%, 8.90\%, and 23.16\%, respectively, with the pruning rate at 30\% and two clusters.

In comparison with Hermes, SAFL exhibits gains of 0.56\% and 2.65\% in inference accuracy on non-IID CIFAR-10 and MNIST, respectively, while using the same number of clusters (five and two, respectively). 
Both methods utilize models of the same size. 
Unlike Hermes, which learns model structure information independently, SAFL leverages the structural insights shared among similar clients, significantly enhancing sub-model performance.
When the data labels within each cluster are consistent, the accuracy reaches its peak. As the number of clusters decreases, 
there is a slight decline in accuracy. However, even when the cluster number is reduced to one, SAFL still outperforms Hermes. 
This indicates that despite variations in clients' data distributions, there remains valuable structural information that can be leveraged to enhance inference accuracy.

\subsection{Comparison with Hermes with Different Pruning Rates}

Table~\ref{tab:table3} presents the performance analysis of SAFL and Hermes across various pruning rates for the CIFAR-10 and MNIST datasets. 
The results unequivocally demonstrate that SAFL consistently outperforms Hermes, showcasing robustness even at higher levels of model compression.

For CIFAR-10, at a pruning rate of 0.3, SAFL shows a clear advantage with an accuracy of 93.29\%, compared to 92.73\% for Hermes. 
As the pruning rate increases, the performance gap widens. 
Notably, at the most aggressive pruning rate of 0.7, while Hermes experiences a significant decrease in accuracy of 12.09\% from its initial rate of 0.3, SAFL maintains a respectable accuracy of 89.85\%, only a 4.40\% reduction from its initial rate, sustaining a lead of over 9\% against Hermes.

The MNIST dataset exhibits a similar trend, with SAFL surpassing Hermes at all evaluated pruning rates. 
At a pruning rate of 0.3, SAFL's accuracy stands at 93.29\%, versus Hermes' 90.65\%. 
At a pruning rate of 0.65, SAFL achieves an accuracy of 91.54\%, which is only a 1.75\% drop from its rate at 0.3, continuing to outperform Hermes, whose accuracy drops to 86.78\%, a 3.87\% decline from its rate at 0.3.

SAFL's architecture not only demonstrates superior performance over Hermes, particularly in managing non-IID data distributions typical of federated learning environments, 
but also illustrates the effectiveness of SAFL in balancing trade-offs between model size and accuracy, crucial in scenarios demanding significant reductions in model footprint.
These results underscore that, in contrast to Hermes's approach of pruning based solely on local data, SAFL's strategy of learning structural information from other similar clients significantly enhances performance. 
This becomes increasingly apparent as the pruning rate escalates, clearly delineating the advantages of SAFL's methodology.

\subsection{Inference Accuracy vs. Communication Cost}

Figure~\ref{fig eight} compares SAFL with the baselines in inference accuracy and communication cost.
Communication costs are normalized such that the cost for FedAvg is set to 1.0. The normalized communication costs for other methods are then calculated relative to this baseline.
Since the model employed in FedAvg does not include BN layers, the communication costs for FedAvg and FedBN are equivalent.
When using the Cifar-10 dataset, we set the pruning rate of SAFL and Hermes to 0.4, and when using the MNIST dataset, we set the pruning rate of SAFL and Hermes to 0.6. 
For the Cifar-10 dataset, SAFL has a higher inference accuracy than other baselines while only spending 40.06\% of the communication overhead of FedAvg. 
For the MNIST dataset, SAFL only spends 33.01\% of the communication overhead of FedAvg. 
Compared with Hermes, SAFL introduces a clustering mechanism, and the client and server need to transmit cluster model information additionally, which adds a little extra overhead, which is 0.86\% in CIFAR-10 and 0.62\% in MNIST. 
However, a small amount of additional communication overhead is worthwhile in exchange for the improvement in accuracy.
Overall, these results affirm the robustness and communication efficiency of SAFL, making it a preferable choice in federated learning scenarios where minimizing communication costs without compromising accuracy is critical.

\section{Conclusion}
In this paper, we introduced SAFL, a novel federated learning framework that effectively integrates iterative clustering and pruning techniques. 
This innovative approach allows each client to efficiently learn structural information from peers possessing similar data characteristics, thereby facilitating the identification and enhancement of their ideal sub-models. This method effectively tackles the challenges associated with heterogeneous data distributions prevalent in federated learning environments.
Our extensive experimental evaluation confirms the effectiveness of SAFL in markedly reducing model size while simultaneously improving inference accuracy. These attributes position SAFL as a highly beneficial framework for federated learning deployments, especially in settings where data heterogeneity and model efficiency are critical concerns.
In summary, SAFL not only offers a robust approach for navigating complex and varied data environments in federated systems but also establishes a new standard for future investigations into efficient and personalized federated learning solutions.

\newpage

\bibliographystyle{plain} 


\vfill

\end{document}